\documentclass[10pt,twocolumn,letterpaper]{article}
\pdfoutput=1

\usepackage{ijcb}
\usepackage{times}
\usepackage{epsfig}

\usepackage{subcaption}
\usepackage{placeins}
\usepackage{url}
\usepackage{booktabs}
\usepackage{amsmath}
\usepackage{amssymb}
\usepackage{graphicx}
\usepackage{xcolor}

\usepackage[linesnumbered,ruled]{algorithm2e}
\usepackage[utf8]{inputenc}
\usepackage[russian,english]{babel}

\makeatletter
\let\NAT@parse\undefined
\makeatother
\usepackage[sort,numbers]{natbib}

\graphicspath{ {images/} }

\ijcbfinalcopy 

\ifijcbfinal\fi

\makeatletter
\def\ps@IEEEtitlepagestyle{
\def\@oddfoot{\mycopyrightnotice}
\def\@evenfoot{}
}
\def\mycopyrightnotice{
{\hfill \footnotesize 978-1-7281-9186-7/20/\$31.00 \copyright 2020 IEEE\hfill}
}
\makeatother

\begin{document}

\title{Long-Term Face Tracking for Crowded Video-Surveillance Scenarios}

\author{Germán Barquero, Carles Fernández and Isabelle Hupont\\
Herta\\
C/ Pau Claris 165, 4B - 08037 Barcelona (Spain)\\
{\tt\small \{german.barquero,carles.fernandez,isabelle.hupont\}@hertasecurity.com}
}

\maketitle
\thispagestyle{empty}

\begin{abstract}
Most current multi-object trackers focus on short-term tracking, and are based on deep and complex systems that do not operate in real-time, often making them impractical for video-surveillance. In this paper, we present a long-term multi-face tracking architecture conceived for working in crowded contexts, particularly unconstrained in terms of movement and occlusions, and where the face is often the only visible part of the person. Our system benefits from advances in the fields of face detection and face recognition to achieve long-term tracking. It follows a tracking-by-detection approach, combining a fast short-term visual tracker with a novel online tracklet reconnection strategy grounded on face verification. Additionally, a correction module is included to correct past track assignments with no extra computational cost. We present a series of experiments introducing novel, specialized metrics for the evaluation of long-term tracking capabilities and a video dataset that we publicly release. Findings demonstrate that, in this context, our approach allows to obtain up to 50\% longer tracks than state-of-the-art deep learning trackers.  
\end{abstract}

\let\thefootnote\relax\footnotetext{\mycopyrightnotice}

\FloatBarrier
\section{Introduction}

\begin{figure}[t!]
\centering
\includegraphics[width=\linewidth]{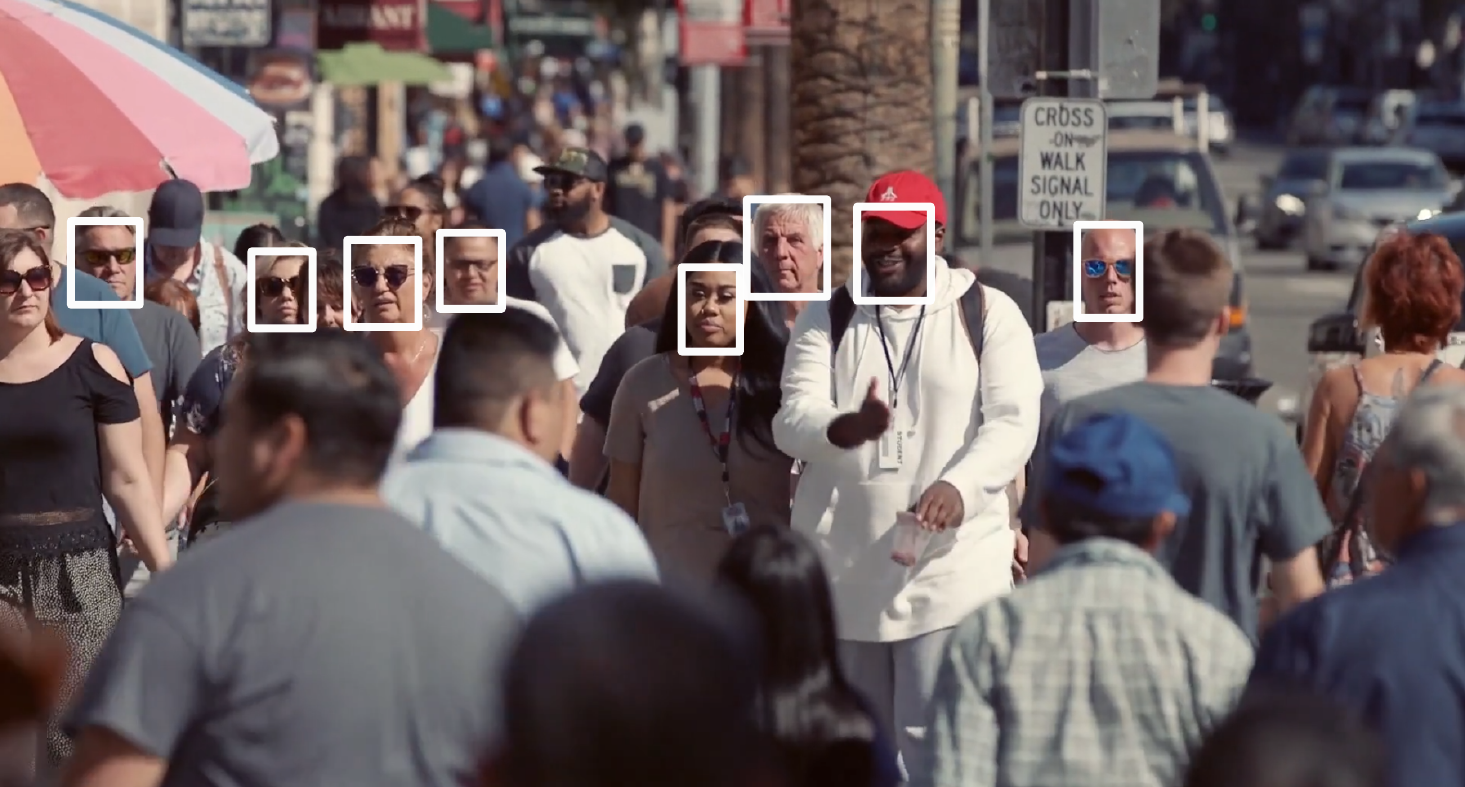}
\caption{Example of crowded video-surveillance scenario targeted in this work. Faces (white bounding boxes) are most of the time the only visible part of the subjects to be tracked. In this scenario, the proposed face tracking approach surpasses general object tracking methods.}
\label{fig:crowded_onlyfaces}
\end{figure}

Recent advances on Convolutional Neural Networks (CNNs) and improvements on IP cameras (e.g. new lenses, 4K/8K resolutions) have allowed to move video-surveillance systems to increasingly crowded, large-scale and unconstrained scenarios.    
These novel scenarios typically involve crowds of people massively walking toward cameras located at near eye level, as illustrated in Figure~\ref{fig:crowded_onlyfaces}. 

When a face recognition system is deployed, alarms are sent to end-users (e.g. police bodies) every time a new subject is detected or identified. The subject is generally tracked, and one single alarm is generated per track to avoid sending multiple duplicate alarms. Obtaining long tracks is therefore of the utmost importance to increase the system's usability.  Moreover, a subject may remain on scene for seconds or even minutes. Making enforcement bodies wait until the end of tracks to receive alerts (which is known as \textit{offline} strategy) implies losing a precious time that could save lives. Hence, real-time and online constraints become essential. However, in this context, it is particularly challenging to perform reliable online long-term tracking of subjects, since: 

\begin{itemize}
    \item The face is often the only visible part of the person. Consequently, the tracking algorithm has to rely on the facial region exclusively, and not on more extended full-body or upper-body regions as in classic pedestrian tracking works \cite{li2014datasetCUHK,chen2019integrated}.
    \item People in video-surveilled places typically move all around the scene, positioning themselves closer or farther from the camera focus, and becoming occluded or blurred for long periods. Existing generic object trackers cannot handle these situations properly, as demonstrated in \cite{lin2019mobiface}.
    \item There is a lack of available datasets covering this kind of scenarios. Crowded video-surveillance videos are particularly difficult to collect and annotate.
\end{itemize}

In this work, we propose an architecture especially conceived for long-term face tracking in crowded video-surveillance contexts. More specifically, our main contributions can be summarized as follows:

\begin{itemize}
    \item Our architecture recovers from partial and full long-term occlusions thanks to a novel online tracklet reconnection module grounded on face verification techniques.
    \item We also propose a track correction module, which updates past track assignments with current information. This module has no extra computational cost, while considerably improving long-term tracking performance.
    \item We validate the system with regard to different state-of-the-art trackers, and present an ablation study quantifying the contribution of each proposed module. Four validation metrics designed to evaluate long-term tracking capabilities are introduced for that purpose.
    \item We publicly release to the community the video dataset used in our experiments, including ground truth track annotations.
\end{itemize}

\section{Related work}

\subsection{Multi-object tracking}

Multi-object tracking is commonly carried out by assigning a single-object tracker to  each target of interest. Current state-of-the-art single-object trackers are based on Deep Learning (DL). A number of approaches rely on Siamese CNNs to discriminate the regions proposed by a region proposal network \cite{zhu2018dasiamrpn, li2019siamrpn++, wang2019siammask, voigtlaender2019trackingsegm}. Other works suggest saving a pool of templates from past track images, the most representative and different among them, and use them to match regions of interest \cite{sauer2019holistic}. Nevertheless, a common drawback of these DL-based methods is that they are computationally expensive and cannot handle long-term tracking: they fail to re-locate out-of-view targets when re-appearing in the scene (see \cite{lin2019mobiface} for a comprehensive survey). A longer-term DL tracker has been very recently proposed in \cite{wangtracking}, but it requires an initial offline training of the model with images from the particular target to track, and thus it is not suitable for our use case.

More computationally efficient trackers are not based on DL, but also achieve competitive accuracies. In \cite{bochinski2018viou}, the use of a Kernelized Correlation Filter (KCF) visual tracker is proposed to fill the gaps generated after applying a classic intersection-over-union (IOU) data association strategy between frame detections. This algorithm works at high frame rates, but still relies heavily on IOU values, which makes it prone to ID-switches. Other high-performing visual trackers include: MOSSE~\cite{bolme2010visual} and CSRT~\cite{lukezic2017discriminative}, which are both based on correlation filters; and the Median Flow tracker \cite{kalal2010forward}, based on motion flow. However, again, these trackers tend to fail with long-term occlusions.

\subsection{Multi-face tracking}

The tracking of persons has been mainly tackled by applying generic object tracking approaches and targeting full-body regions. Few studies rely exclusively on faces to track persons and address face tracking as a problem with its own particularities. 

Taking advantage of the high accuracies reached by current face detectors, many face tracking works propose tracking-by-detection approaches. In \cite{comaschi2015online}, a generic AdaBoost face detector is combined with an adaptive structured output SVM tracker, using a IOU data association strategy. However, this approach is only suitable for short-term tracking, as it does not implement any tracklet reconnection strategy. 
As a longer-term approach, \cite{kalal2010face} applies the Tracking-Learning-Detection (TLD) paradigm to faces: the face is tracked and simultaneously learned by a detector that supports the tracker once it fails. 

More recent approaches achieve long-term face tracking by using clustering techniques to make associations between short-term tracklets \cite{lin2018offline, zhang2016offline}. Short-term tracklets are obtained by combining detectors and simple data association methods. Then, facial features are computed for each detection through DL face recognition models, and clusters are extracted from the feature space to collapse same-identity tracklets. Although these approaches achieve state-of-the-art results, they work fully offline and imply a high computational cost, which is not suitable for real-time tracking.

It is also worth mentioning that some works propose tracking mechanisms to improve face recognition in videos, in scenarios where persons of interest are previously enrolled using few still images. 
In \cite{dewan2016adaptive} and \cite{zheng2018automatic}, simple visual trackers are used to obtain tracks from which new (unseen) high-quality face stills are collected \cite{dewan2016adaptive}. These stills are matched against reference images to identify people. Interestingly, they are additionally used to enrich the gallery of enrolled images, thus improving face recognition performance. 
Nevertheless, these works are devoted to face recognition, and face tracking is just treated as a secondary task.

\subsection{Datasets for face tracking in crowds}
\label{sec:public_datasets}

Studies on people tracking have traditionally focused on full-body pedestrian tracking in low-to-moderately crowded urban scenes. As a result, several pedestrian video datasets are available and commonly used by the community \cite{li2014datasetCUHK, liu2018datasetShanghaiTech, deng2014datasetPETA}. Another field for which a large number of datasets is available is crowd analysis, e.g. crowd counting or crowd behavior understanding~\cite{rodriguez2011datasetDriven, zhang2015datasetWorldExpo, dendorfer2019cvpr19}. These crowd datasets usually contain high-angle views, in which people faces appear at very low resolutions (mostly below 30x30 pixels). Very recently, the MobiFace dataset has been released to evaluate in-the-wild face tracking algorithms for mobile devices \cite{lin2019mobiface}. Videos are recorded from moving smartphone cameras, sometimes in ``selfie'' mode, and contain few faces (less than 5) per video. Consequently, none of these datasets cover our use case.

The only exception is the ChokePoint dataset. It provides a collection of 48 videos capturing individual subjects walking through two portals \cite{wong2011chokepoint}. To pose a more challenging real-world surveillance problem, two extra sequences were recorded in a indoor crowded environment, which represent the scenarios that we are targeting in this work. 

\section{Proposed tracking architecture}

\begin{figure*}[ht]
\centering
\includegraphics[width=16cm]{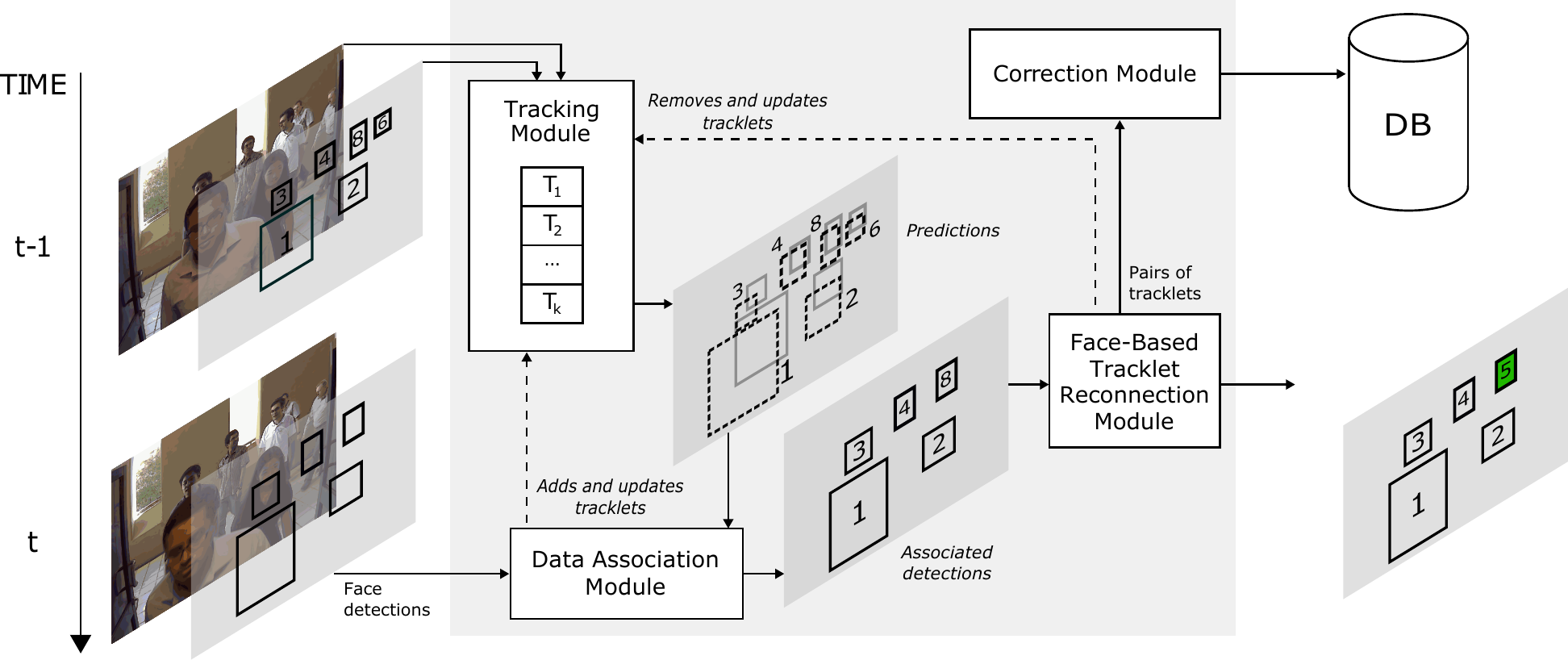}
\caption{Overview of the proposed four-module tracking architecture.}
\label{fig:system_scheme}
\end{figure*}

This work presents a four-module architecture that overcomes the limitations of previous approaches to favor long-term face tracking in crowded video-surveillance environments, see Figure \ref{fig:system_scheme}.   
The following sections describe each module in detail. 

\subsection{Tracking module (TM)}
\label{sec:STFT}

The system firstly extracts tracklets following a tracking-by-detection approach.
The tracking module is in charge of predicting face locations over the frames where the face detector fails, using one simple visual tracker. Several visual tracking algorithms are available for that purpose in our implementation, including KCF, MOSSE, Median Flow and CSRT.  

The tracking module creates a tracklet ${T_i}$ for every new detected face $i$. In case the detector loses a face in the following frame, it keeps predicting its position until the data association module decides to: (i) update the position with a new detection or (ii) force it to die. These tracklets are additionally used to collect the pool of reference images that serve as a basis for the face identification mechanism, c.f. Section \ref{sec:reID}.

\subsection{Data association module (DA)}
\label{sec:DA}

In order to decide which detection should guide which tracklet, a data association problem needs to be solved. 
Once the tracking module has predicted the new $K_t$ positions of live tracklets for a frame $t$, the data association module retrieves the $N_t$ faces detected in that frame. A state-of-the-art face detector is used in our implementation for that purpose \cite{zhang2017faceboxes}. Then, to establish correspondences between predicted and detected face bounding boxes, the Munkres implementation of the Hungarian algorithm is applied to their IOU values \cite{kuhn1955hungarian}. For every correspondence with an IOU value above a threshold $\lambda_{IOU}$, the tracking module updates the corresponding tracklet with the new detected bounding box position. For detected faces without a tracklet correspondence, a new tracklet is initialized and these faces are considered as new identities.

Tracklet predictions without a face detection correspondence are kept alive for a maximum number of frames $T_{max}$. The tracking module keeps predicting their location over those frames where no association is made. If the tracklet is not updated for $T_{max}$ frames consecutively, it is forced to die and marked as inactive.

    
    

\subsection{Face-based tracklet reconnection module (FBTR)}
\label{sec:reID}

When a partial or full occlusion occurs, trackers generally lose the tracked target and consider it as a new object when it re-appears. To overcome this limitation, our system incorporates an online face-based tracklet reconnection module (FBTR). This module is inspired by face verification: a face recognition model is used to collect reference image templates from each tracklet, and then a matching procedure is applied to unify same-identity tracklets. In our implementation, we use the state-of-the-art face recognition model by \cite{deng2019arcface}.

The selection of reference templates is driven by image quality. More particularly, three indicators are considered: (i) face detection confidence, (ii) head pose angles and (iii) a blur metric.
Detection confidence is a value directly provided by the face detector. Pitch, yaw and roll head pose angles are estimated using the fiducial extractor by Zhu et al. \cite{zhu20173ddfa}. The blur quality metric is obtained by applying the Laplace operator in the facial area as proposed by Nikitin et al. \cite{nikitin2014facequality}.

\begin{figure}[b!]
\centering
\includegraphics[width=\linewidth]{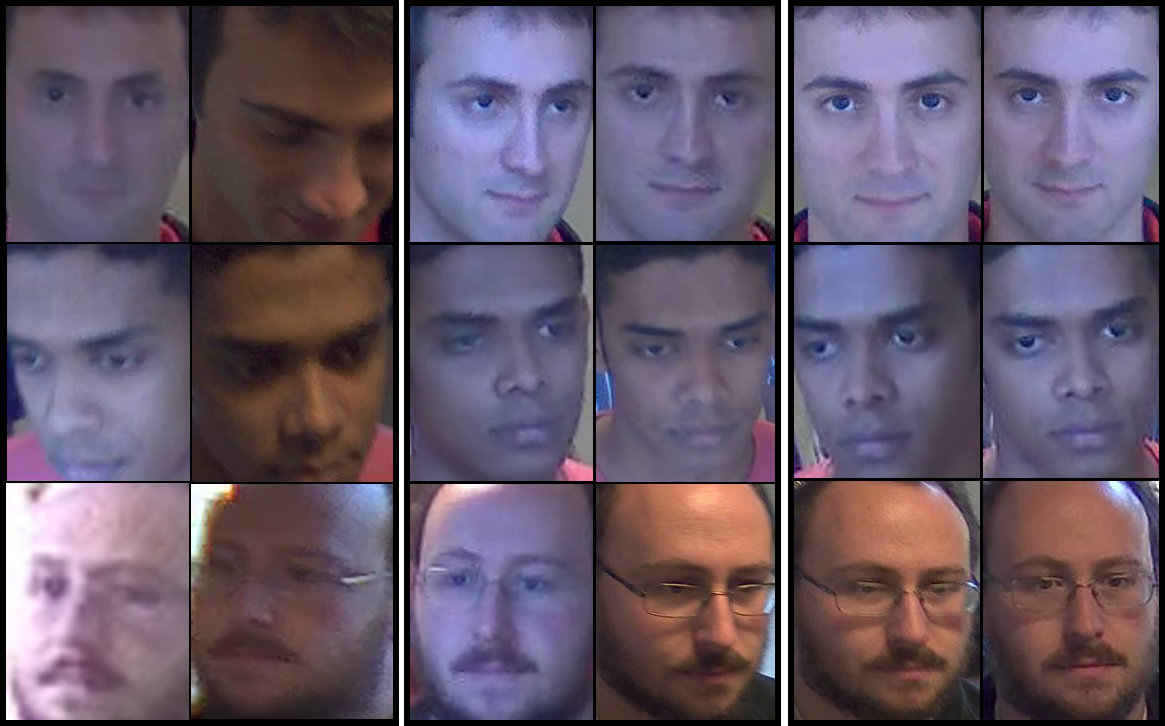}
\caption{Examples of discarded (left), verifiable (center) and enrollable (right) faces.}
\label{fig:sample_snaps}
\end{figure}

Using these quality indicators, face detections contained in each tracklet are divided into 3 groups (see Figure \ref{fig:sample_snaps}):

\begin{itemize}
\item \textit{Enrollable faces.} Faces with high visual quality (detection confidence \textgreater0.95, head angles in range $\pm25^{\circ}$ and blur quality \textgreater0.9). They are used to enroll identities.
\item \textit{Verifiable faces.} Faces that have enough quality to extract a reliable template from them (detection confidence \textgreater0.8, head angles in range $\pm60^\circ$ and blur quality between 0.75-0.9). In the tracklet reconnection process, verifiable faces are matched against enrollable faces. Note that enrollable faces are a subset of verifiable faces.
\item \textit{Discarded faces.} Their low quality makes them unsuitable for the FBTR module, as they would provide non-reliable templates.
\end{itemize}

After the data association phase, the FBTR component checks the quality of each detected face. If not discarded, a template is extracted with the face recognition model and stored either as an enrollable or verifiable one. Then, for each tracklet $T_k$ with an assigned detection $D_k$ in the current frame, we retrieve tracklets $T_i$ with $i\neq k$. For each tracklet $T_i$, the mean of its enrollable face templates, $\overline{E_{T_i}}$, is computed and taken as the tracklet reference template. The mean of the verifiable face templates of $T_k$, $\overline{V_{T_k}}$, is also computed. Now, considering $S$ the similarity function of the face recognition model, the tracklet $T_i$ with which $T_k$ is identified must verify:

\vspace{2mm}

$\underset{{\{T_i\}}}{\arg\max}\hspace{1mm}S(\overline{E_{T_i}}, \overline{V_{T_k}})\hspace{2mm}\text{subject to}\hspace{1.2mm} 
 S(\overline{E_{T_i}}, \overline{V_{T_k}})\geq \lambda_{FBTR}$
 
 \vspace{3mm}
 
\noindent Where $\lambda_{FBTR}$ is an identification score threshold. If any $T_i$ verifies the previous condition, the FBTR module re-assigns detection $D_k$ from tracklet $T_k$ to $T_i$. Tracklets $T_k$ and $T_i$ are joined and the pair $\langle T_k, T_i\rangle$ is appended to a list of track pairs, which is the input to the correction module.

\begin{figure}[t]
\includegraphics[width=\linewidth]{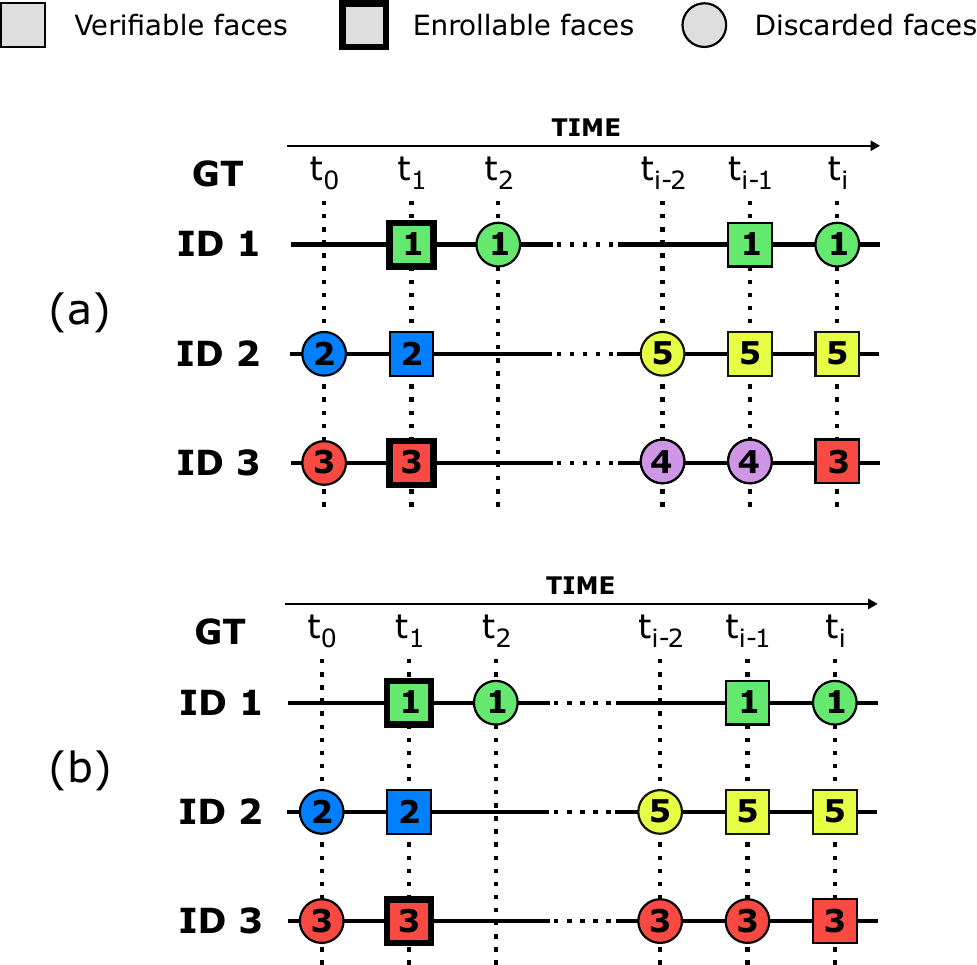}

\caption{Results of our tracking approach without (a) and with (b) the correction module. Numbers inside detections correspond to the tracklet identifier assigned by our tracker. In (b), the identification at frame $t_{i}$ updates prior frames.}
\label{fig:reid_tracklets}
\end{figure}

Figure \ref{fig:reid_tracklets}a illustrates the tracklet reconnection procedure. Tracklet~$4$ is reconnected to $3$ as soon as a verifiable face is found, while tracklet~$5$ cannot be reconnected to tracklet $2$ because no enrolment face was available.

\subsection{Correction module (CM)}

The correction module receives the list of tracklet pairs provided by the FBTR module. For each pair $\langle T_k, T_i\rangle$, it retrieves all the detections assigned in the past to $T_k$ and switches their track ID to $T_i$. Figure \ref{fig:reid_tracklets}b shows the outcome of adding this module. Note that the correction module now replaces the whole tracklet~$4$ by tracklet~$3$ at frame $t_i$.

The benefit of this module is that it refines the tracking process without adding any extra computational cost. It is of particular interest for forensic video analysis.

\section{Evaluation metrics}


In this section, we introduce novel metrics especially suitable for evaluating the long-term tracking performance of our system. These metrics are grounded on those commonly used in multi-object tracking, such as mismatch errors (number of ID changes produced across tracks) or track fragmentations \cite{milan2016mot16}.

One of the main challenges in multi-object tracking is to avoid drifting, i.e. losing a target. Assuming that the facial detector has high accuracy, in our use-case drifting only happens when two or more tracklets switch their target identity. The drifting effect becomes much more dramatic when incorporating reconnection capabilities: when switching its target, the tracklet integrity is lost, leaving the system vulnerable to future misassignments. Therefore, there is a need for revisiting the mismatch error metric by taking into account two new important concepts:

\begin{itemize}
    \item \textit{Soft-mismatch error (smme).} It is produced when the tracker switches the correct identity (ID) to a new one that has not been associated to any track until that time. This leads to ID-fragmentation, but can potentially be recovered by the FBTR module.
    \item \textit{Hard-mismatch error (hmme).} It is produced when, instead of switching to a new identity, the tracker switches to a previously assigned one. This leads to a probably unrecoverable ID-switch.
\end{itemize}

Another desired feature of our system is its ability to obtain long-term tracklets. Thus, it is necessary to introduce metrics able to quantify the length of tracklets. Overall, the goal is to build a robust long-term tracker that reduces fragmentation while keeping the number of ID-switches low. To achieve it, we formulate three new scalar metrics and a graphical one: \\

\noindent\textbf{Fragmentation (Frag)} 
\begin{equation*}
    Frag=\frac{\sum{smme}}{\#dets}
\end{equation*}
where the numerator is the sum of soft-mismatch errors produced in a video, and $\#dets$ is the total number of faces detected (according to ground truth annotations).\\

\noindent\textbf{ID-Switches (IDSW)}
\begin{equation*}
    IDSW=\frac{\sum{hmme}}{\#dets}
\end{equation*}
where $hmme$ is the total number of hard-mismatch errors in the video.\\

\noindent\textbf{Completion Rate Plot (CRP)}. 
Plot showing the percentage of tracks (vertical axis) that had at least $X\%$ of their ground truth detections correctly identified (horizontal axis). In other words, it plots $CR_{X}$ values for $0\leq X \leq 100$, where:

\begin{equation*}
    CR_X=\frac{\textit{\# IDs correctly tracked for at least X\% of the time}}{\textit{total number of IDs}}
\end{equation*}

\noindent\textbf{Completion Rate Sum (CRS)}. 
Area under the curve of the completion rate plot. The higher this value is, the longer subjects have been successfully tracked.
\begin{equation*}
    CRS=\frac{\sum_{X=1}^{100}{CR_X}}{100}\hspace{0.5cm}
\end{equation*}


\section{Experiments and results}

This section describes the dataset used to evaluate the proposed architecture, and the results of the different experiments carried out with it. All experiments were conducted on a desktop Intel i7-7700 CPU machine at 3.60GHz and a GeForce GTX 950 GPU, and implemented in Python. The thresholds of the system, $\lambda_{IOU}$ and $\lambda_{FBTR}$, were set based on a private collection of training videos to 0.25 and 0.7, respectively.

\subsection{Evaluation dataset}

\begin{table*}[h!]
    \centering
    \small
    \begin{tabular}{ccccccccc}
        \toprule
         & Resolution & Length & Face dets. & Density & Scenario & \#Subjects & Description \\
         \midrule
        Choke1 & 800x600 & 1'24" & 7964 & 4.0 & Indoor & 24 (*) & Corridor recorded from 3 cameras over a door.\\
        Choke2 & 800x600 & 1'11" & 8710 & 4.8 & Indoor & 26 (*) & Corridor recorded from 3 cameras over a door.\vspace{1mm}\\
        Street & 1920x1080 & 1'8" & 4883 & 2.9 & Outdoor & 31 & Street scene filmed from a low-eye level angle.\\
        Sidewalk & 1920x1080 & 27" & 8433 & 13.0 & Outdoor & 34 &  Crowd walking to the camera, filmed at eye level.\\
        Bengal & 1920x1080 & 40" & 6953 & 6.9 & Outdoor & 36 & A pedestrian scene filmed at eye level.\\
         \bottomrule
    \end{tabular}
    \caption{Description of the videos used to evaluate our tracking architecture. In videos marked with asterisk (*), subjects leave and re-enter the scene twice. Density refers to the mean number of face detections per frame.}
    \label{tab:videos}
\end{table*}

Since there are no public datasets fully corresponding to our use case (c.f. Section \ref{sec:public_datasets}), we have compiled and annotated a set of 5 videos showing  crowded indoor and outdoor video-surveillance scenes. 

Two of these videos come from the extra sequences (cameras P2E\_S5 and P2L\_S5) of the well-known public dataset ChokePoint \cite{wong2011chokepoint}. To force re-appearances of subjects and validate the FBTR module performance, the three sequences recorded by P2E\_S5 cameras were concatenated, leading to the \textit{Choke1} video. Similarly, the video \textit{Choke2} was generated from \textit{P2L\_S5} cameras sequences. The remaining three videos were selected from YouTube. Table \ref{tab:videos} details each video content and properties.

Tracks were semi-automatically annotated to obtain ground truth data. Firstly, face bounding boxes were retrieved using a state-of-the-art face detector \cite{zhang2017faceboxes}. Only faces with a detection confidence above 0.50 were considered. Then, detections were manually verified, and tracks and corresponding IDs were annotated\footnote{All videos and annotations are publicly available at \url{https://github.com/hertasecurity/LTFT}.}.

\subsection{Benchmarking of face trackers} 
\label{sec:STFT_benchmark}

This first experiment aims at benchmarking different trackers to choose the most appropriate one for the tracking module. We analyze the performance of our data association module using the following visual trackers: CSRT~\cite{lukezic2017discriminative}, KCF~\cite{bochinski2018viou}, Median Flow~\cite{kalal2010forward} and MOSSE~\cite{bolme2010visual}. Additionally, we include in our benchmark the longer-term tracker TLD \cite{kalal2010face} and the state-of-the-art deep tracker SiamRPN++~\cite{li2019siamrpn++}. 

\begin{figure}[h]
    \centering
    \includegraphics[width=\linewidth]{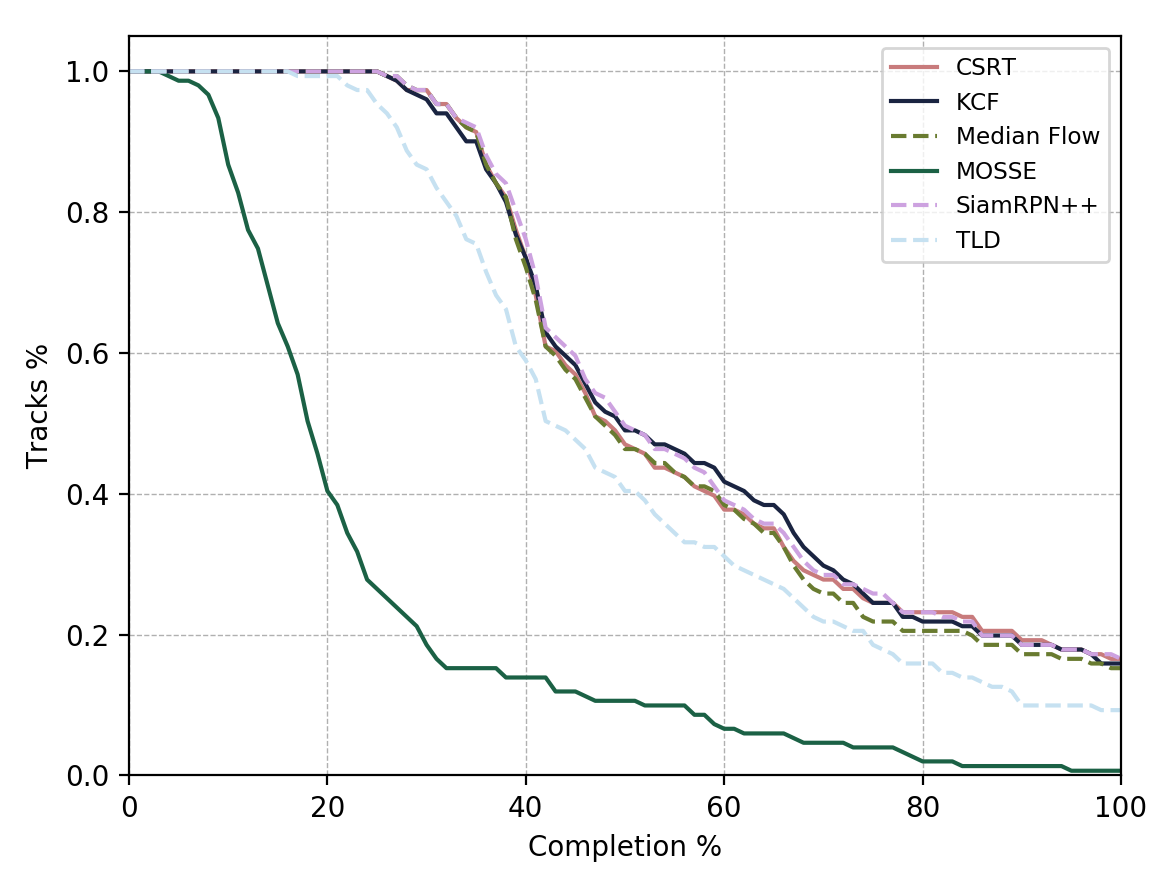}
    \caption{CRP for each tracker tested in the tracking module.}
    \label{fig:stft_CRP}
\end{figure}

Table~\ref{tab:results_STFT} and Figure~\ref{fig:stft_CRP} present the results obtained on all of our evaluation videos. Results highlight the lower performance of TLD. This is probably due to its capability to re-identify targets, which makes it vulnerable to ID-switches. The best performing tracker in terms of ID-switches is MOSSE, but its high fragmentation leads to a poor overall completion rate. The SiamRPN++ tracker achieves the highest completion rate ($CRS=0.590$) and the lowest fragmentation ($Frag=0.01264$). However, KCF is very close to it ($CRS=0.587$, $Frag=0.01299$) and runs at much higher FPS, which is critical in video-surveillance contexts. 

According to these findings, simple visual trackers and more sophisticated DL-based trackers lead to similar performances in such challenging environments where occlusions are extremely frequent. In the following experiments we will use the KCF tracker as the core of our tracking module.

\begin{table}[h]
    \centering
    \small
    \begin{tabular}{ccccc}
        \toprule
        Tracker & Frag & ID-Switches & CRS & FPS\\
        \midrule
        CSRT & 0.01326 & 0.00379 & 0.581 & 3.80\\
        KCF & 0.01299 & 0.00395 & \textbf{0.587} & \textbf{22.12}\\
        Median Flow & 0.01470 & 0.00352 & 0.574 & 16.26\\
        MOSSE & 0.21831 & \textbf{0.00192} & 0.234 & 21.63\\
        SiamRPN++ & \textbf{0.01264} & 0.00420 & \textbf{0.590} & 1.74\\
        TLD & 0.03711 & 0.00912 & 0.515 & 1.51\\
        \bottomrule
    \end{tabular}
    \caption{Evaluation of the data association module on the whole dataset, using different trackers in the tracking module.}
    \label{tab:results_STFT}
\end{table}

\subsection{Ablation study}

In this section, we present an ablation study that quantifies the contribution of each module (TM, DA, FBTR and CM) and demonstrates the suitability of the proposed approach for long-term tracking in crowded video-surveillance scenarios.

Five ablation experiments are presented: 

\begin{itemize}
    \item \textbf{DA.} Tracking is performed following a simple data association strategy: the tracking module is deactivated, and data association is computed based on the IOU value of detections in consecutive frames.
    \item \textbf{DA+FBTR.} Same as DA, plus face identification.
    \item \textbf{DA+TM.} In this experiment, data association is carried out using KFC in the tracking module. No face identification or correction mechanisms are applied.
    \item \textbf{DA+TM+FBTR.} Same as DA+TM, plus FBTR.
    \item \textbf{DA+TM+FBTR+CM.} Same as DA+TM+FBTR, plus correction capabilities.
\end{itemize}

\begin{figure}[t]
    \centering
    \includegraphics[width=\linewidth]{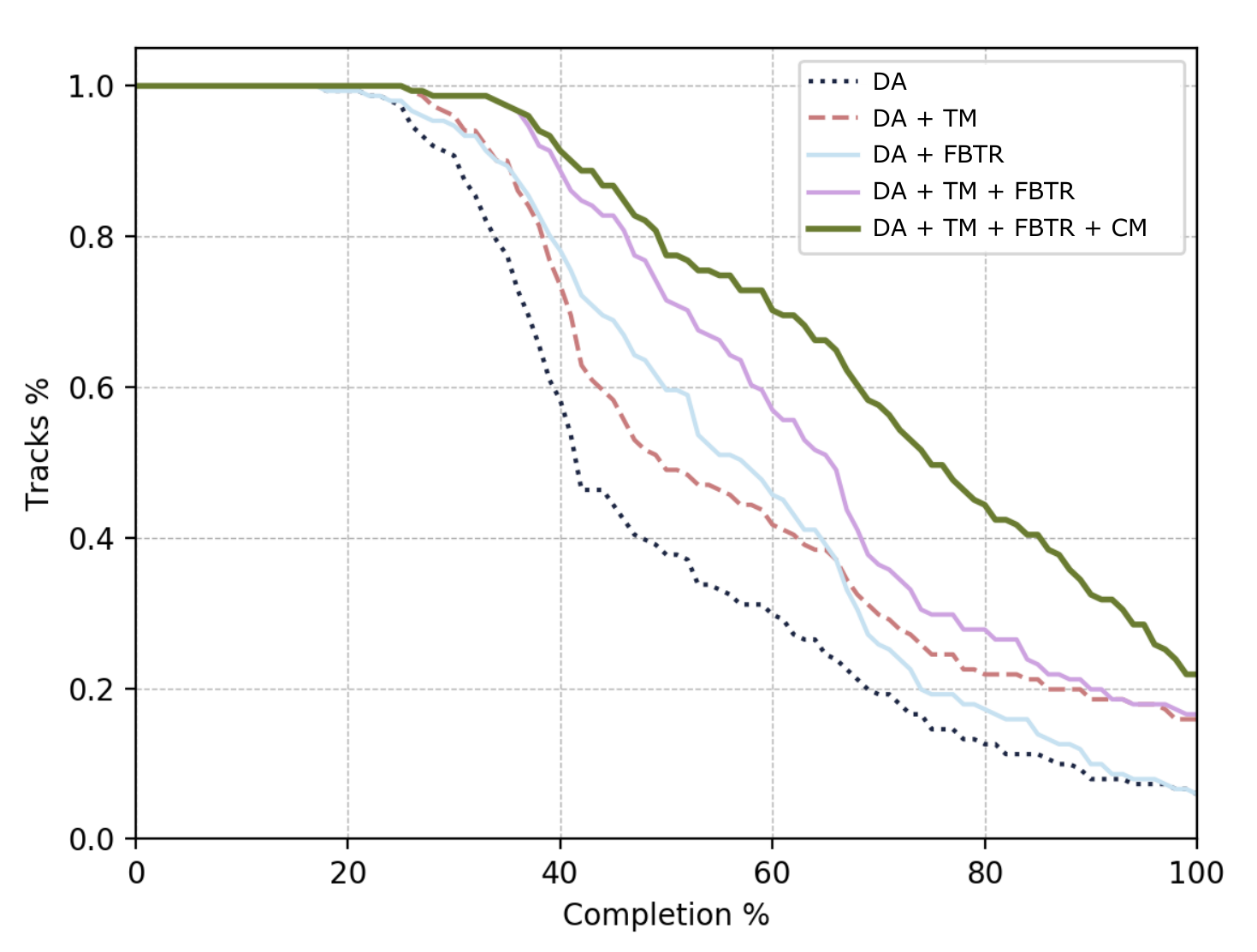}
    \caption{CRPs for the different ablation experiments.}
    \label{fig:ablation_CRP}
\end{figure}

Table \ref{tab:ablation_ALL_table} and Figure \ref{fig:ablation_CRP} show the results of the ablation study on our whole dataset. It can be clearly observed that each added module increases the overall tracking completion rate. The impact of FBTR is particularly noteworthy: it increases the CRS by a 15.5\% (from 0.503 to 0.581) and a 12.2\% (from 0.587 to 0.659) when added to DA and DA+TM, respectively.
The correction module also improves long-term tracking (CRS increase of 11.5\%, from 0.659 to 0.735) without any extra computational cost. At the same time, it strongly reduces the fragmentation generated by the FBTR module by a 22.8\% (from 0.1573 to 0.1215). As expected, the number of ID-Switches increases as we achieve longer-term tracking, but its value stays reasonably low (0.00357 when FBTR is used).

Table \ref{tab:ablation_tables} details ablation results per video. 
Results on \textit{Choke1} and \textit{Choke2} highlight the good performance of our architecture, especially of the FBTR module (CRS increases above 50\%), in contexts where people leave and re-enter the scene. The impact of the CM is also dramatic, reaching CRS values up to 0.845 and reducing fragmentation up to 40\%.
The remaining videos do not contain subject re-appearances. In the case of  \textit{Sidewalk}, where long occlusions are frequent, FBTR and CM improve short-term tracking by a 7.3\% (CRS=0.718) and  11.2\% (CRS=0.744), respectively.
\textit{Street} and \textit{Bengal} are the most challenging videos in terms of illumination, motion and occlusions. In these videos, the impact of FBTR and CM is lower, but the DA+TM+FBTR+CM architecture is still the most successful one.

\begin{table}[t]
    \centering
    \small
    \begin{tabular}{lcccc}
        \toprule
        Tracking architecture & Frag & ID-Switches & CRS & FPS\\
        \midrule
        DA & 0.02677 & \textbf{0.00097} & 0.503 & \textbf{30.0}\\
        DA+FBTR & 0.02940 & 0.00114 & 0.581 & 12.9\\
        DA+TM & 0.01299 & 0.00395 & 0.587 & 22.1\\
        DA+TM+FBTR & 0.01573 & 0.00357 & 0.659 & 10.9\\
        DA+TM+FBTR+CM & \textbf{0.01215} & 0.00357 & \textbf{0.735} & 10.9\\
        \bottomrule
    \end{tabular}
    \caption{Results of the ablation study on our whole dataset.}
    \label{tab:ablation_ALL_table}
\end{table}

\begin{table}[t]
    \centering
    \small
    \begin{tabular}{lccccc}
        \toprule
        Choke1 (800x600) & Frag & ID-Switches & CRS & FPS\\
        \midrule
        DA & 0.01607 & \textbf{0.00001} & 0.375 & \textbf{50.4}\\
        DA+FBTR & 0.02022 & \textbf{0.00001} & 0.572 & 20.1\\
        DA+TM & 0.01017 & 0.00013 & 0.397 & 39.0\\
        DA+TM+FBTR & 0.01406 & 0.00013 & 0.605 & 17.1\\
        DA+TM+FBTR+CM & \textbf{0.00829} & 0.00013 & \textbf{0.720} & 17.1\\
        \midrule
        
        Choke2 (800x600) & & & &\\
        \midrule
        DA & 0.03387 & \textbf{0.00069} & 0.355 & \textbf{53.3}\\
        DA+FBTR & 0.03869 & 0.00103 & 0.511 & 18.1\\
        DA+TM & 0.01584 & 0.00253 & 0.400 & 34.3\\
        DA+TM+FBTR & 0.02124 & 0.00241 & 0.558 & 14.8\\
        DA+TM+FBTR+CM & \textbf{0.01447} & 0.00253 & \textbf{0.845} & 14.8\\
        \midrule
        
        Sidewalk (1920x1080) & & & &\\
        \midrule
        DA & 0.02929 & \textbf{0.00237} & 0.547 & \textbf{15.2}\\
        DA+FBTR & 0.03154 & 0.00261 & 0.618 & 4.7\\
        DA+TM & 0.00996 & 0.00984 & 0.669 & 9.2\\
        DA+TM+FBTR & 0.01150 & 0.00901 & 0.718 & 3.9\\
        DA+TM+FBTR+CM & \textbf{0.00949} & 0.00889 & \textbf{0.744} & 3.9\\
        \midrule
        
        Street (1920x1080) & & & &\\
        \midrule
        DA & 0.03031 & \textbf{0.00123} & 0.580 & \textbf{14.9}\\
        DA+FBTR & 0.03051 & 0.00143 & 0.577 & 13.8\\
        DA+TM & \textbf{0.01884} & 0.00410 & 0.634 & 15.6\\
        DA+TM+FBTR & 0.01946 & 0.00348 & 0.633 & 12.2\\
        DA+TM+FBTR+CM & \textbf{0.01925} & 0.00348 & \textbf{0.637} & 12.2\\
        \midrule
        
        Bengal (1920x1080) & & & &\\
        \midrule
        DA & 0.02459 & \textbf{0.00058} & 0.587 & \textbf{16.2}\\
        DA+FBTR & 0.02488 & \textbf{0.00058} & 0.607 & 7.8\\
        DA+TM & 0.01222 & 0.00288 & 0.732 & 12.5\\
        DA+TM+FBTR & 0.01323 & 0.00244 & 0.735 & 6.6\\
        DA+TM+FBTR+CM & \textbf{0.01194} & 0.00244 & \textbf{0.741} & 6.6\\
        \bottomrule
    \end{tabular}
    \caption{Results of ablation experiments per video.}
    \label{tab:ablation_tables}
\end{table}

\section{Conclusions}

In this work, we have presented an architecture for long-term multi-face tracking in crowded video-surveillance scenarios. The proposed method benefits from the advances in the fields of face detection and face recognition to achieve long-term tracking in contexts particularly unconstrained in terms of movement, re-appearances and occlusions. 

We have introduced specialized metrics conceived to evaluate long-term tracking capabilities, and publicly released a dataset with videos representing the targeted use case. The series of experiments carried out with them lead to interesting findings. Firstly, we show that in such challenging contexts, state-of-the-art deep trackers have a similar performance to other simpler and computationally faster visual trackers. Secondly, we demonstrate that our novel FBTR strategy, which is grounded on face verification, allows to obtain up to 50\% longer tracks. Finally, the proposed cost-free correction module has been proved to increase tracking robustness, not only by keeping on improving long-term capabilities, but also by reducing fragmentation.


\section{ACKNOWLEDGMENTS}
This work was partly funded by the Spanish project AI-MARS (CIEN CDTI Programme, grant number IDI-20181108). I. Hupont was supported by the Torres Quevedo Programme (PTQ-16-08735).

{\small
\bibliographystyle{ieee}
\bibliography{bibliography}
}

\end{document}